\documentclass{article}

\PassOptionsToPackage{numbers,sort&compress}{natbib}
\usepackage[dblblindworkshop,final]{neurips_2026}
\workshoptitle{NEmo: Neuro-Symbolic Embodied Intelligence}

\usepackage[utf8]{inputenc}
\usepackage[T1]{fontenc}
\usepackage{amsmath,amssymb}
\usepackage{graphicx}
\usepackage{wrapfig}
\usepackage{tikz}
\usetikzlibrary{arrows.meta,positioning,calc}
\usepackage{booktabs}
\usepackage{algorithm}
\usepackage{algorithmic}
\usepackage{listings}
\usepackage{fvextra}
\usepackage{microtype}
\usepackage{xcolor}
\usepackage{stmaryrd}
\usepackage[hidelinks]{hyperref}
\usepackage{url}

\graphicspath{{figures/}}
\hypersetup{pdfauthor={},pdftitle={Iterative Policy Refinement through Semantic Rollout Analysis}}

\title{Iterative Policy Structure Refinement through Semantic Rollout Analysis}
\author{
    Feiyu Gavin Zhu$^{1}$ \quad
    Qi Xu$^{2}$ \quad
    Zhifei Deng$^{2}$ \quad
    Zhigang Hua$^{2}$ \quad
    Luke Simon$^{2}$ \\[0.1em]
    \bf Jean Oh$^{1}$ \quad
    \bf Reid Simmons$^{1}$ \\[0.5em]
    $^{1}$Carnegie Mellon University \quad
    $^{2}$Meta \\
    \texttt{feiyuz@cs.andrew.edu}
}

\begin{document}
\maketitle

\begin{abstract}
Structured policies improve efficiency, robustness, and interpretability in imitation learning by introducing task-specific inductive bias, but existing structure generation methods rely either on extensive human input or on static domain knowledge encoded in LLMs, which may be inconsistent with the expert demonstrations.
We propose a closed-loop framework that iteratively refines structured policies using LLM-guided analysis of policy rollouts.
By logging rollouts as semantically meaningful tabular data and prompting the LLM to generate diagnostic analysis code, our method identifies suboptimalities in the policy structure and iteratively corrects them without requiring human instruction.
Experiments on car racing and door opening tasks show that our approach improves imitation learning performance by up to 15\% over zero-shot LLM-generated structures and requires 75\% less compute to achieve the same reinforcement learning performance.
These results demonstrate that tabular rollout analysis provides an effective feedback signal to align LLM-generated policy structures with expert demonstrations, and we can utilize it to generate good policy structures automatically.
\end{abstract}

\section{Introduction}
\label{sec:introduction}

Black-box deep learning models are highly non-linear and can exhibit unpredictable malicious behaviors \cite{betley2025weird} or inconsistent performance \cite{luo2025beyond}.
Recent work addresses this by using LLMs to generate task-specific policy structures that are sparse and semantically meaningful \cite{zhu2025sample}, improving the efficiency and robustness of imitation learning.
However, these approaches rely heavily on human experts to provide domain knowledge or to continuously monitor and refine the generated structures.

In principle, LLMs already encode substantial task-relevant knowledge from pretraining and could be used to instantiate policy structures.
However, because LLMs do not directly observe the environment, the generated structures may encode modeling assumptions that conflict with expert demonstrations or overlook critical variable interactions.
For example, an LLM may introduce a damping term that links steering to angular velocity to correct spins.
However, in demonstrations, an expert may steer in the same direction as the angular velocity during sharp turns, causing the learned policy to violate the intended assumption and exhibit unwanted oscillatory behavior.

\begin{figure}[!t]
\centering
\includegraphics[width=\linewidth,trim=0bp 69bp 0bp 127bp,clip]{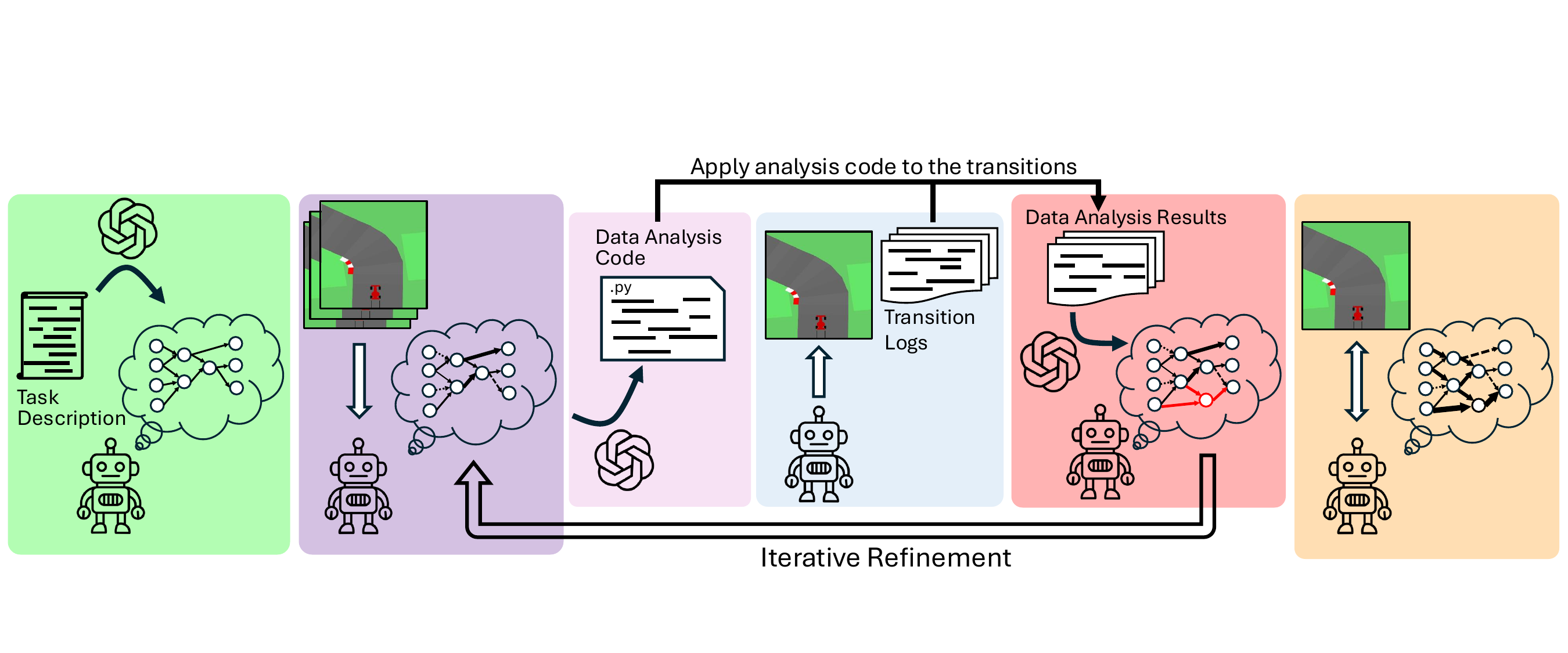}
\caption{Overview of policy refinement.
\textbf{1. Generate the initial structure} by querying an LLM with the task description.
\textbf{2. Update parameters} through imitation learning on the demonstrations.
\textbf{3. Generate analysis code} based on the desired behavior of the policy and its structure topology.
\textbf{4. Collect tabular trajectories}, recording observations, actions, and named latent values during policy rollouts.
\textbf{5. Execute analyses and revise the structure} based on the analysis results.
The updated program returns to parameter fitting and another refinement cycle.
\textbf{6. Reinforcement learning} after the structure aligns with the demonstrations to go beyond the demonstration performance.
}
\label{fig:overview}
\end{figure}

To resolve such misalignment, we propose a closed-loop framework that alternates demonstration-based parameter fitting with LLM-guided structural refinement (Figure~\ref{fig:overview}).
The process resembles software debugging: a programmer specifies expected behavior, writes tests, and uses their results to locate and correct errors.
Here, the LLM generates data analysis code to test expected policy behavior, runs it on tabular rollouts, and uses the results to revise the structure before fitting its parameters again.
This debugging interface is possible because structured policies are sparse and semantically meaningful: each intermediate quantity has an interpretable role, and its dependencies can be traced to specific computations that the LLM can inspect and edit.
For the turning example above, tests can examine whether the angular-velocity term damps rotation as intended and whether steering oscillates during turns.
The results can guide updates to that term or its parameter constraints, such as using \texttt{softplus} to enforce an intended coefficient sign.
Unlike prior methods that rely on visual \cite{chen2024elemental} or textual feedback \cite{peng2024adaptive}, our approach executes analysis code outside the LLM on complete rollout tables and returns only the results, allowing it to handle trajectories beyond the LLM's context limit and data that are not easily visualizable.

Our contributions are threefold: (1) LLM-driven automated policy structure generation and refinement without human expert input, (2) the use of tabular rollout analysis as environment feedback for policy updates, and (3) empirical evidence that structure refinement improves the efficiency of reinforcement learning from demonstrations in control tasks.

\section{Related work}
\label{sec:related}

\subsection{Structured policies and learning from demonstrations}
\label{sec:related-structured}

Since embodied policies can often be naturally expressed in code, many works have explored incorporating LLM code generation into policy representations.
The most direct approaches generate executable policies, symbolic plans, or production rules from language instructions~\citep{liang2022code,liu2023llm+,kirk2024improving}.
Other approaches generate intermediate components that define behavior indirectly, such as spatial value maps and language-guided state abstractions~\citep{huang2023voxposer,peng2024learning,peng2024adaptive}.
More recent graph-based approaches, including Graph-as-Policy (GaP), compose perception, planning, and control modules into interpretable computation graphs~\citep{chen2026gap}.

Demonstration-guided approaches use expert knowledge, narrated actions, or interactive corrections to construct interpretable representations of behavior~\citep{zhu2025sample,kim2025interactiveprogramsynthesismodeling,zhu2026interpret}.
Our representation follows KIM~\citep{zhu2025sample}: generated code specifies a sparse semantic structure, while continuous parameters are learned from demonstrations.
We extend this separation by generating tabular data analysis code to identify misalignment and guide structural updates, reducing the need for human corrective instructions.

Reinforcement learning from demonstrations improves learning efficiency by shaping rewards, augmenting replay data, or initializing policies and dynamics models~\citep{brys2015reinforcement,hester2018deep,uchendu2023jump,hansen2022modem}.
Our work uses demonstrations to fit parameters within successively refined structures, so subsequent reinforcement learning starts with both a learned policy and a demonstration-aligned structure.

\subsection{LLM-guided program revision and execution feedback}
\label{sec:related-refinement}

Classical neural architecture search automates structure design through differentiable, evolutionary, or learned search over model primitives~\citep{liu2018darts,jian2023eenas,zoph2018learning}.
LLM-guided methods instead use semantic knowledge and execution feedback to revise programs directly, including self-debugging, algorithm evolution, and coding-agent improvement~\citep{chen2023selfdebug,novikov2025alphaevolve,zhang2025darwin}.
Related methods revise reward functions or state representations using learning and environment feedback~\citep{ma2023eureka,li2025R,peng2024adaptive}.

In robotics, programmatic search and agentic learning systems use rollout feedback to improve control programs, reusable skills, or the surrounding learning workflow~\citep{hu2025multimodal,lu2026aspire,xiao2026enpire}.
Graph-based refinement extends this process to modular policy graphs, with GaP using simulation rehearsal to improve graph structure and parameters~\citep{chen2026gap}.

Our pipeline uses code generation in two complementary ways: one program defines the trainable policy structure, and another analyzes its tabular rollouts.
The sparse semantic structure connects numerical feedback to specific computations that can be revised, while demonstrations fit the continuous parameters between revisions.
This differs from searching only over complete executable policies or optimizing a reward function: the analysis code inspects the intermediate values of the policy itself to guide structural refinement.

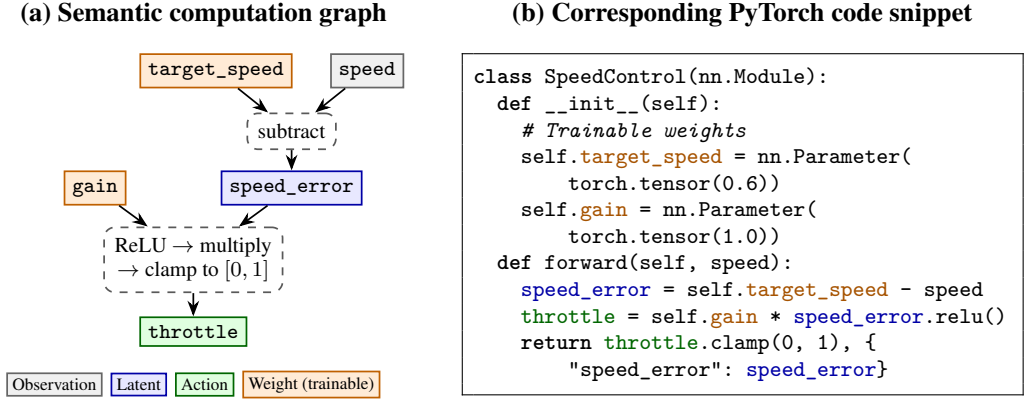
\begin{figure}[!t]
\centering
\begin{minipage}[t]{.47\linewidth}
\vspace{0pt}\centering
\textbf{(a) Semantic computation graph}\\[9pt]
\begin{tikzpicture}[
    x=1cm,y=1cm,>=Stealth,
    var/.style={draw,line width=.65pt,inner sep=3pt,font=\ttfamily\fontsize{8}{9}\selectfont},
    obs/.style={var,draw=black!60,fill=black!6},
    latent/.style={var,draw=blue!65!black,fill=blue!9},
    action/.style={var,draw=green!40!black,fill=green!12},
    weight/.style={var,draw=orange!75!black,fill=orange!16},
    op/.style={draw=black!65,dashed,rounded corners=4pt,
        line width=.65pt,fill=white,inner sep=4pt,font=\fontsize{8}{9}\selectfont,align=center},
    flow/.style={->,line width=.65pt},
    key/.style={font=\fontsize{6.5}{8}\selectfont,inner sep=2pt}
]
\node[weight] (target) at (0,0) {target\_speed};
\node[obs] (speed) at (2,0) {speed};
\node[op] (subtract) at (1,-.8) {subtract};
\node[latent] (error) at (1,-1.55) {speed\_error};
\node[weight] (gain) at (-1.6,-1.55) {gain};
\node[op] (control) at (-.3,-2.5) {ReLU $\to$ multiply\\$\to$ clamp to $[0,1]$};
\node[action] (throttle) at (-.3,-3.45) {throttle};
\draw[flow] (target) -- node[left,font=\scriptsize] {} (subtract);
\draw[flow] (speed) -- node[right,font=\scriptsize] {} (subtract);
\draw[flow] (subtract) -- (error);
\draw[flow] (error) -- (control);
\draw[flow] (gain) -- (control);
\draw[flow] (control) -- (throttle);
\matrix[column sep=3pt] at (-.3,-4.15) {
    \node[obs,key] {Observation}; &
    \node[latent,key] {Latent}; &
    \node[action,key] {Action}; &
    \node[weight,key] {Weight (trainable)}; \\
};
\end{tikzpicture}
\end{minipage}\hfill
\begin{minipage}[t]{.51\linewidth}
\vspace{0pt}\centering
\textbf{(b) Corresponding PyTorch code snippet}\\[5pt]
\begin{lstlisting}[language=Python,firstline=4,
    basicstyle=\ttfamily\fontsize{8.5}{10}\selectfont,
    emph={target_speed,gain},emphstyle=\color{orange!65!black},
    emph={[2]speed_error},emphstyle={[2]\color{blue!65!black}},
    emph={[3]throttle},emphstyle={[3]\color{green!40!black}}
]
class SpeedControl(nn.Module):
  def __init__(self):
    # Trainable weights
    self.target_speed = nn.Parameter(
        torch.tensor(0.6))
    self.gain = nn.Parameter(
        torch.tensor(1.0))
  def forward(self, speed):
    speed_error = self.target_speed - speed
    throttle = self.gain * speed_error.relu()
    return throttle.clamp(0, 1), {
        "speed_error": speed_error}
\end{lstlisting}
\end{minipage}
\caption{
A sample speed-control subgraph and its corresponding PyTorch implementation.
Solid nodes distinguish observations (gray), semantic latents (blue), action outputs (green), and trainable weights (orange); dashed nodes denote operation nodes (some collapsed due to space constraints).
The returned \texttt{speed\_error} exposes the latent value for further rollout analysis.
}
\label{fig:topology}
\end{figure}

\section{Semantic policy structure}
\label{sec:structure}
\label{sec:policy-program}
\label{sec:running-example}
\label{sec:semantic-interface}

Following previous work \cite{zhu2025sample}, a structured policy computes actions from observations through a semantically meaningful latent space.
Formally, each structured policy has a symbolic component $\llbracket\alpha\rrbracket = \langle (V, P), E \rangle$: a bipartite graph whose variable nodes $V$ represent intermediate values, operator nodes $P$ compute transformations between them, and edges $E$ connect variables and operators.
Each structured policy also has a neural component $\Theta$ containing all learnable latent values (a subset of $V$), which determine the exact continuous output.
We implement each structured policy as a task-specific PyTorch model \cite{paszke2019pytorch}, representing the graph topology in code and fitting its parameters through the autograd function.
We denote the policy with structure $\llbracket \alpha \rrbracket$ and trained weights $\Theta$ by $\pi_{\langle \llbracket \alpha \rrbracket, \Theta \rangle}$, using $\llbracket \alpha \rrbracket$ for both the graph topology and its PyTorch implementation.

Figure~\ref{fig:topology} shows a minimal speed-control example alongside its underlying implementation.
The topology defines a proportional controller for maintaining constant speed, while imitation learning fits parameters such as \texttt{target\_speed} and \texttt{gain} to the demonstration's speed and responsiveness.
This highlights the separation of symbolic and neural components: the symbolic structure is task-specific, specifying the intermediate steps that generate actions; continuous parameters are demonstration-specific, determining how the demonstration completes the task.

Unlike post-hoc explanations of black-box models, latent semantics directly reveal expected policy behavior, while intermediate computations trace how actions are generated.
We audit all the internal mechanisms that generate actions and use this information to detect structure-demonstration misalignment and propose structural refinement.
\section{Iterative refinement through semantic analysis}
\label{sec:method}

We draw inspiration from software development: programmers write tests for intermediate steps to flag potential issues and use the feedback to improve the software. We treat the structured policy similarly to develop through this iterative refinement process (Figure~\ref{fig:overview}).

\subsection{Generating the initial policy}
\label{sec:initial-generation}

We generate $\llbracket \alpha^{(0)} \rrbracket$ by querying the LLM with only the task description, whereas prior work requires demonstrator-provided knowledge of how demonstrations are generated.
We use the superscript $(i)$ to denote the policy after $i$ iterations.
Relying on pretrained knowledge and an incomplete environment description, the LLM typically generates simple structures (e.g., Figure~\ref{fig:latent-traces}(a)) that may misalign with the structure underlying the demonstration.
Subsequent refinement aims to close this structure-demonstration gap.

\subsection{Fitting policy parameters from demonstrations}
\label{sec:fitting}

Given the policy structure, we still need to tune its parameters to complete the policy. We use imitation learning to maximize the probability of demonstration actions given demonstration states.
Formally, given a demonstration as a set of state-action pairs,
\begin{equation}
    \mathcal{D} = \{ \langle s_0, a_0 \rangle, \langle s_1, a_1 \rangle, \cdots \}
\end{equation}
we train the weights $\Theta^{(i)}$ of structure $\llbracket \alpha^{(i)} \rrbracket$ with the negative log-likelihood loss \cite{osa2018algorithmic}
\begin{equation}
    \Theta^{(i)} = \arg\max_\Theta \mathbb{E}_{s, a \in \mathcal{D}} [- \log P\big(a \mid \pi_{\langle \llbracket\alpha^{(i)}\rrbracket, \Theta\rangle}(s)\big)]
\end{equation}

\begin{figure}[!t]
\centering
\begin{minipage}[t]{.485\linewidth}
\centering\textbf{\small (a) Before: fixed centering strength}\\[3pt]
\resizebox{\linewidth}{!}{\input{figures/diagrams/incremental_before}}
\end{minipage}\hfill
\begin{minipage}[t]{.485\linewidth}
\centering\textbf{\small (b) After: centering amplified when off-track}\\[3pt]
\resizebox{\linewidth}{!}{\input{figures/diagrams/incremental_after}}
\end{minipage}
\par\vspace{5pt}
\begin{minipage}[t]{.485\linewidth}
\centering\textbf{\small (c) Recovery behavior analysis}\\[6pt]
\raggedright\fontsize{8}{10}\selectfont
\textbf{Expected Behavior:} \\
When the race car is far from the track center,\\
it should steer back toward the center.\\[5pt]
\textbf{Analysis Code:} \\
\begin{lstlisting}[basicstyle=\ttfamily\fontsize{7}{8.5}\selectfont,frame=single,aboveskip=0pt,belowskip=4pt]
# Filter: far from the track center
off_track = abs(log.offset) > half_track_width
frames = log[off_track]
# Compare: steering toward center?
toward = frames.steering * frames.offset > 0
fraction = mean(toward)
\end{lstlisting} 
\textbf{Analysis Result:} \\
Only \textcolor{red!80!black}{\textbf{14.9\%}} satisfy the check\\
\end{minipage}\hfill
\begin{minipage}[t]{.485\linewidth}
\centering\textbf{\small (d) Recorded lateral offset and steering}\\[3pt]
\includegraphics[width=\linewidth]{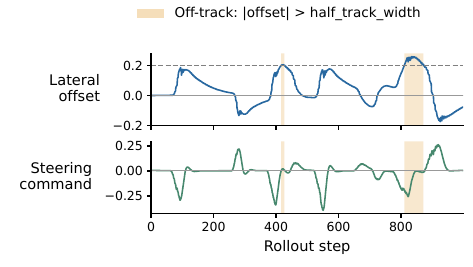}
\end{minipage}
\caption{Example of off-track recovery structure refinement.
(a) The steering structure before refinement (other nodes omitted);
(b) The refined structure adds stronger steering when the race car is off-track;
(c) The desired behavior of re-centering, corresponding test, and the result;
(d) Recorded lateral offset and steering.
}
\label{fig:latent-traces}
\end{figure}

In Figure~\ref{fig:latent-traces}(a), imitation learning fits \texttt{centering\_weight} to determine how quickly the policy steers toward the track center.
This highlights the division of labor: the LLM knows that a race car generally needs to stay near the track center and encodes this knowledge in the structure. However, it does not know how aggressively the car should recenter, so this information must be learned from demonstrations and reflected in the weights while keeping the structure unchanged.

\subsection{Generating analysis code from the policy structure}
\label{sec:diagnostics}

By capturing semantics, the policy structure provides an interface for detailed inspection and localizing issues in each subgraph.
As when generating unit tests in software development, we instruct the LLM to reflect on expected behaviors and generate corresponding tests.
Given the task description and current structure $\llbracket \alpha^{(i)} \rrbracket$, the LLM proposes hypotheses and writes Python analysis functions $T^{(i)}_j$ to test each of the hypotheses.

Figure~\ref{fig:latent-traces}(c) illustrates one hypothesis: the policy should attempt to recenter after the car goes off-track.
The policy structure connects this hypothesis to a specific editable computation: the centering contribution generated from lateral offset (subgraph shown in Figure \ref{fig:latent-traces} (a)).
The test directly queries relevant variables (e.g., \texttt{offset}). We take advantage of the flexibility of Python syntax and have no limit on what kind of tests it can generate.
The above example filters off-track time steps and measures the percentage of time steps in which the policy exhibits recovery behavior.

As these tests specify general desired behavior, they can be generated directly from the policy structure, independently of actual rollouts.
It shows a unique advantage: exposing semantic latents lets the LLM test specific questions, such as whether a steering contribution sufficiently influences off-track behavior, rather than inspecting only the aggregate action.
During each iteration, the LLM generates multiple tests, each testing an individual behavior and subgraph of the policy.
Other generated analyses examine action saturation, reconstruction errors, and speed regulation (Appendix~\ref{app:diagnostic-prompts}).

\subsection{Collecting tabular trajectories from policy rollouts}
\label{sec:traces}

After training $\pi_{\langle \alpha^{(i)}, \Theta^{(i)} \rangle}$, we execute it and collect trajectories for analysis.
At each time step, the policy takes in the current state $s_t$ and populates all the latent values $z^{(i)}_{0,t}, z^{(i)}_{1,t}, \dots, z^{(i)}_{k,t}$ in its computation graph $\llbracket \alpha^{(i)} \rrbracket$.
Therefore, in addition to the output distribution $a_t$, it also returns all of the intermediate latent states, such as $\texttt{lateral\_offset}_t = 0.1$.
We organize observations, latents, parameters, and actions into a tabular trajectory,
\begin{equation}
    X^{(i)}[t,:]=[s_t,z^{(i)}_{0,t}, z^{(i)}_{1,t}, \dots, z^{(i)}_{k,t},\Theta^{(i)},a_t,s_{t+1}]
    \label{eq:trace}
\end{equation}
where each row records the current state, intermediate latents, weights, action output, and next state.

Figure~\ref{fig:latent-traces}(d) plots these two recorded columns over the full 1,000-step rollout.
The traces show the raw values supplied to the recovery behavior code analysis.
This shows the advantage of a semantic latent space: recorded trajectories contain not only states and actions but also the physical meaning of the intermediate computations that generate each action.

\subsection{Executing the analysis code and revising the policy structure}
\label{sec:revision}

We execute tests $T^{(i)}$ for desired behaviors on rollout $X^{(i)}$ through a Python interpreter and collect their outputs.
The tests can be applied to arbitrarily long rollouts or large rollout tables without exceeding the LLM's context length. Their results indicate whether the policy meets each criterion and, if not, which part of its structure is suboptimal.

In Figure~\ref{fig:latent-traces}(c), the test finds center-directed steering in only $14.9\%$ of off-track steps.
It identifies a state-dependent discrepancy: the executed policy rarely selects center-directed steering where the analysis expects recovery when the race car is off the track. This motivates a structural update making the steering more aggressive when the race car is off the track.

The LLM revises the structure using the task description, current structure, analysis code, and analysis outputs.
Edits can change dependencies, operations, latent variables, coefficient constraints, and initialization.
Figure~\ref{fig:latent-traces}(b) shows the recovery branch after the revision.
The revised program adds a pathway explicitly handling off-track behavior.
It computes an off-track indicator measuring excess lateral offset, scales it by a trainable nonnegative recovery strength, and adds one to form a multiplier.
The additional operations make recovery strength depend explicitly on the state while retaining the other steering contributions.
The program exposes the new indicator as \texttt{latent\_offtrack\_amount}, allowing subsequent data analysis code to inspect when this branch is active.

The illustrated change creates a \textbf{structural update}, rather than merely changing a fitted parameter.
Multiple analysis scripts may expose several issues in the structured policy during each update iteration, and the LLM is instructed to address all of them in that iteration.
The revised structure then repeats the cycle of training, expected-behavior test generation, rollout inspection, and revision.

\subsection{Reinforcement learning on the demonstration-aligned structure}
\label{sec:rl-method}

Convergence in the policy structure $\llbracket \alpha \rrbracket$ indicates no structural improvement is needed for the policy.
To further improve performance, we apply a finite-length variant of GSPO \citep{zheng2025group} reinforcement learning to further optimize the continuous parameters $\Theta$.
Specifically, given $k$ trajectories per step, we use a sliding window of length $l$ to break each of them into tracklets of length $l$, then form groups such that the starting positions of the groups are similar.
Appendix~\ref{app:rl} describes the implementation details.
\section{Experiments and analysis}
\label{sec:experiments}

We evaluate iterative refinement on Car Racing and Door Opening using one demonstration per task.
The experiments test whether iteration improves policies, compare the complete pipeline with gradient learning and program evolution, and examine how semantic analysis guides refinement.

\subsection{Environment setup and choice of baselines}
\label{sec:setup}

We choose two tasks with complementary challenges: Car Racing tests long-horizon behavior where small errors accumulate, while Door Opening requires distinct approach and pull stages with abrupt changes in appropriate actions at contact.
Both tasks use continuous actions and low-dimensional semantic observations (details in Appendix~\ref{app:environment}).
A human using a joystick provides the Car Racing demonstration, and a motion planner provides the Door Opening demonstration.

We compare with the following three baselines:
\begingroup
\setlength{\leftmargini}{1.5em}
\begin{enumerate}
    \item MLP: Unstructured neural policies: a two-layer unstructured MLP that can be learned through both imitation learning and reinforcement learning
    \item Zero-shot Structure: Structured but non-adaptive LLM-generated policies: a structured policy generated by an LLM with the same prompt but without any refinement
    \item MLES \cite{hu2025multimodal}: Fully LLM-driven programmatic evolution: LLM-guided evolutionary search in the policy space; the policy cannot be further trained with demonstrations or rewards.
\end{enumerate}
\endgroup
The refinement experiments use GPT-5-mini with high reasoning effort.
For all imitation and reinforcement learning, we train to convergence.

\begin{table}[!t]
\centering
\small
\setlength{\tabcolsep}{4pt}
\begin{tabular}{@{}lcrrr@{}}
\toprule
\bf Method & \bf Demo & \bf LLM calls $\downarrow$ & \bf Resets $\downarrow$ & \bf Return $\uparrow$ \\
\midrule
Human demonstrations & - & - & - & $871.0 \pm 12.8$ \\
\midrule
MLP + IL & Yes & 0 & 0 & $636.6\pm258.6$ \\
MLP + RL, low compute & Yes & 0 & 400 & $591.8\pm330.5$ \\
MLP + RL, high compute & Yes & 0 & 2000 & $850.2\pm101.4$ \\
Initial structure + IL & Yes & 1 & 0 & $766.5\pm100.4$ \\
Initial structure + RL, low compute & Yes & 1 & 400 & $657.0\pm37.3$ \\
Initial structure + RL, high compute & Yes & 1 & 2000 & $\underline{901.1\pm14.0}$ \\
MLES & No & 2016 & 8064 & $847.1\pm132.5$ \\
MLES, limited queries & No & 16 & 64 & $240.0\pm130.3$ \\
\midrule
Refined structure + IL (ours) & Yes & 9 & 4 & $882.0\pm29.33$ \\
Refined structure + RL, low compute (ours) & Yes & 9 & 404 & $900.5\pm14.2$ \\
Refined structure + RL, high compute (ours) & Yes & 9 & 2004 & $\mathbf{906.5\pm24.0}$ \\
\bottomrule
\end{tabular}
\caption{Car Racing performance and resource usage.
Returns are mean $\pm$ standard deviation over ten random seeds.
Resets include refinement and training interactions, excluding final evaluation.}
\label{tab:results}
\end{table}

\subsection{Iterative refinement and overall performance comparison}
\label{sec:refinement-results}
\label{sec:rl-results}

\begin{wrapfigure}{R}{0.50\textwidth}
\centering
\includegraphics[width=\linewidth]{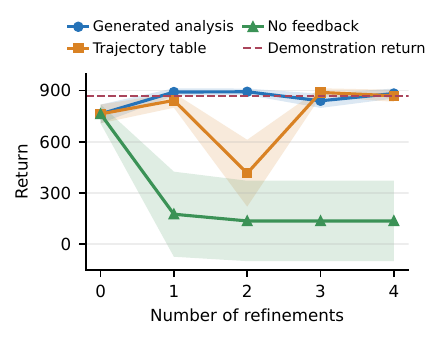}
\caption{Execution feedback improves policy refinement in Car Racing.
Generated analysis avoids the large intermediate drop under direct table feedback.
Shading shows standard deviations.}
\label{fig:result-car}
\end{wrapfigure}

\paragraph{Iteration improves performance}

Table~\ref{tab:results} shows that structural refinement improves performance over the zero-shot structure with no refinement.
Under imitation learning, the refined policy significantly outperforms the zero-shot model by $15\%$ ($p \approx 0.007$), with performance similar to human demonstrations.
This verifies our hypothesis that refinement corrects the mismatch between the LLM's pretrained task knowledge and the demonstration's true structure.

The zero-shot structure requires significantly more RL steps for similar performance ($2000$ vs $404$ resets), showing that a better structure improves RL efficiency.
The zero-shot structure is expected to eventually match performance because it is \textit{inconsistent} with the expert demonstrations, rather than inherently \textit{wrong}.
This also explains why low-budget RL underperforms IL ($p \approx 0.010$): the policy must \textit{unlearn} IL parameters at a local optimum.

\begin{figure}[!t]
\centering
\begin{minipage}[t]{0.24\linewidth}\centering
\includegraphics[width=\linewidth]{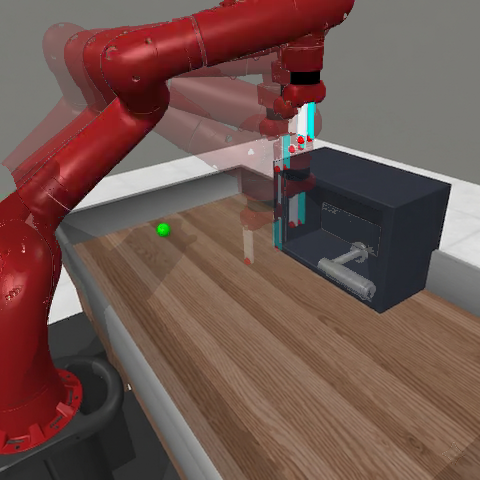}\\
\scriptsize Iteration 0\\$0.548\pm0.0$
\end{minipage}\hfill
\begin{minipage}[t]{0.24\linewidth}\centering
\includegraphics[width=\linewidth]{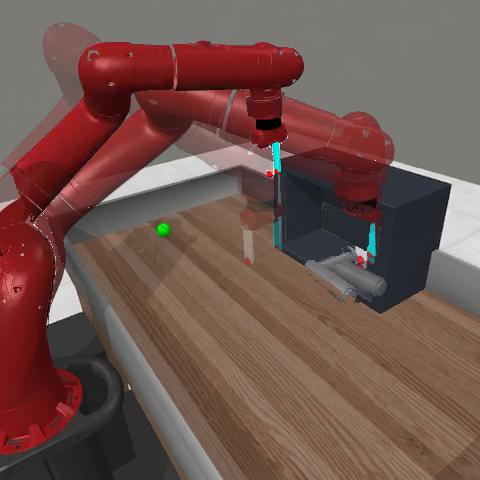}\\
\scriptsize Iteration 1\\$0.545\pm0.008$
\end{minipage}\hfill
\begin{minipage}[t]{0.24\linewidth}\centering
\includegraphics[width=\linewidth]{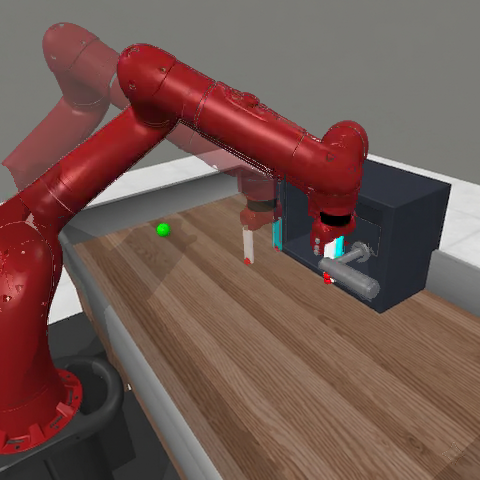}\\
\scriptsize Iteration 2\\$0.540\pm0.023$
\end{minipage}\hfill
\begin{minipage}[t]{0.24\linewidth}\centering
\includegraphics[width=\linewidth]{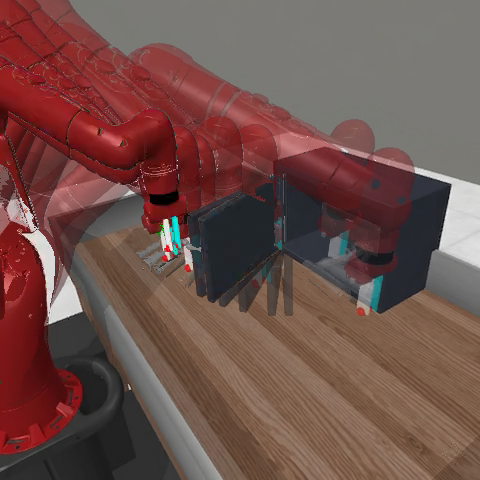}\\
\scriptsize Iteration 3\\$0.110\pm0.059$
\end{minipage}
\caption{Door Opening progresses through three revisions.
The numbers below show the mean and std of distance from handle to target across 10 seeds; lower is better.
}
\label{fig:door_iterations}
\end{figure}

This also holds for Door Opening, where structural refinement produces a reach-and-pull policy that generalizes from one demonstration to unseen configurations (Figure~\ref{fig:door_iterations}).
The initial policy's linear-motion assumption causes upward drift; the next revision adds an arc toward the handle but no pull, before later refinement learns both approach and pull.
Refinement reduces the minimum handle-to-target distance by approximately 80\% ($p<10^{-4}$).

\paragraph{Semantic analysis of refinement.}
\label{sec:edit-analysis}
The generated analyses examine summary statistics, reconstruction errors, correlations and regressions, and temporal behavior to connect policy computations with their observed effects.
These checks guide both changes to existing features and the addition of new interactions: repeated slip checks lead to a revised feature and a sign constraint, while the recovery behavior analysis motivates stronger centering when the car is off-track.
Manual inspection also shows that refinement retains the initial policy's ten strategy categories and adds two more, rather than replacing the controller with a collection of unrelated rules.
Appendix~\ref{app:extended} details the strategy and diagnostic coverage, and Appendix~\ref{app:records} presents the corresponding tests and structural updates.

\paragraph{Performance comparison with other baselines}
The refined structure outperforms the MLP in mean return under each matched learning condition.
After IL, it achieves $38\%$ better performance than MLP ($p\approx0.019$); at the low RL budget, it also achieves $52\%$ better performance ($p\approx0.021$).
At the high RL budget, the refined policy has Table~\ref{tab:results}'s highest mean return (906.5 vs 850.2 for MLP), but the difference is not statistically significant ($p\approx0.136$).
But with RL, the refined structure exceeds the human demonstrations with significance ($p\approx 0.002$), demonstrating that RL enables the policy to perform beyond the demonstrations.

Combining structural refinement with gradient learning also achieves high returns with far fewer LLM queries than program evolution alone.
With 16 queries, MLES reaches 240.0 return, compared with 900.5 for the refined policy with a single demonstration.
MLES reaches similar performance at 847.1 with 2,016 queries, $224x$ the refined policy's query count, and 8,064 environment resets.
It therefore requires a much larger search budget to approach the performance of the refined pipeline.

These comparisons show that combining symbolic structure with continuous updates from demonstrations is more efficient than either component alone.

\begin{wraptable}{R}{0.50\textwidth}
\centering
\small
\setlength{\tabcolsep}{4pt}
\begin{tabular}{@{}lrr@{}}
\toprule
\bf Feedback & \bf \shortstack{Input\\tokens (k) $\downarrow$} & \bf \shortstack{Output\\tokens (k) $\downarrow$} \\
\midrule
Trajectory table & $110.1\pm1.6$ & $17.9\pm1.3$ \\
Analysis results & $26.0\pm3.4$ & $37.5\pm2.9$ \\
\bottomrule
\end{tabular}
\caption{Generated analysis uses fewer input tokens and more output tokens than direct trajectory-table feedback.
Values summarize six refinement rounds, with population standard deviations.}
\label{tab:tokens}
\end{wraptable}

\subsection{Ablation on feedback type}
\label{sec:feedback-results}

Execution feedback improves both the quality and trajectory of policy refinement (Figure~\ref{fig:result-car}).
We compare generated analysis with two conditions: 1) direct trajectory-table feedback, providing the entire rollout table to the LLM, and 2) revision without execution feedback.
Without closed-loop environment feedback, asking the LLM to improve the structure increases misalignment and causes return to collapse.
Both feedback conditions reach high returns, but direct table feedback is more unstable, dropping to 415.4 at the second revision.
The analysis-based condition remains above 839 across all four revisions, providing more consistent improvement along this refinement trajectory.

Our data analysis method uses only $25\%$ of the input tokens required by table feedback (Table~\ref{tab:tokens}).
Output usage doubles because the LLM generates analysis in addition to structure code, but the total number of processed tokens remains much lower.

\section{Limitations}
\label{sec:discussion}
\label{sec:limitations}

Because tests are generated from the existing policy structure and table column names, the LLM may not anticipate every failure mode or generate tests for it. As a result, some problems may never be exposed or corrected.
Additionally, current evaluations cover only low-dimensional tasks following the MDP assumption. A natural next step is to scale the method to more complex tasks with larger feature spaces or possibly a POMDP formulation \cite{kaelbling1998planning}.
The method may also benefit from a fine-tuned LLM and/or a more carefully designed, higher-level domain-specific language such as Scenic \cite{fremont2019scenic}, which may be easier to update than PyTorch.

\section{Conclusion}
\label{sec:conclusion}
\label{sec:takeaway}

We introduced iterative policy structure refinement through semantic rollout analysis.
The method alternates demonstration-based parameter fitting with LLM-generated diagnostics and structural edits, linking execution and revision through semantically meaningful latents.
The sparse, semantic structure allows the LLM to test specific relationships within a policy and use the results to guide structural revision, following the cycle of testing and revision used in code debugging.
With one demonstration per task, refinement improves imitation learning, generalizes to unseen task configurations, and reduces subsequent reinforcement learning resets.
\label{sec:main-end}

\begin{ack}
Feiyu is supported by the SoftBank Group–Arm Fellowship.
This work is part of the Keio-CMU partnership.
\end{ack}

\bibliographystyle{plainnat}
\bibliography{references}


\clearpage
\appendix
\section*{Appendix}
\section{Semantic analysis of policies and diagnostics}
\label{app:extended}

We analyze the strategies represented by the policies and the coverage of generated diagnostics to understand how rollout analysis guides structural refinement.

\subsection{Policy strategy coverage}
\label{app:strategy-coverage}

The policy structure makes both the controller's strategies and the targets of its diagnostics explicit.
We manually group computations in the Car Racing programs into speed-control, steering-control, and special-case strategies (Table~\ref{tab:semantic}).
The initial structure contains ten categories; refinement retains all ten and adds steering toward the outside of curves and braking during sharp turns.
MLES contains fifteen categories, including more rules for grass, curbs, off-track recovery, and stopping.
Our structured policies instead concentrate on general relationships between task state and actions, with specific nuanced behavior handled by continuous parameters.
These categories summarize broad control strategies rather than every operator or structural edit.

\begin{table}[!t]
\centering
\footnotesize
\setlength{\tabcolsep}{4pt}
\begin{tabular}{l|c|c|c}
\toprule
\textbf{Strategy} & \textbf{MLES} & \textbf{Initial} & \textbf{Refined}  \\
\midrule
\multicolumn{4}{l}{\textbf{Speed Control}} \\
\midrule
speed during straights & $\checkmark$ & $\checkmark$ & $\checkmark$ \\
speed at track center & $\checkmark$ & $\checkmark$  & $\checkmark$ \\
slow at corners & $\checkmark$ & $\checkmark$ & $\checkmark$ \\
slow at sides / large offset & $\checkmark$ & $\checkmark$ & $\checkmark$ \\
slow when outside track & $\checkmark$ & & \\
slow at high angular velocity & & $\checkmark$ & $\checkmark$ \\
no concurrent acceleration / braking& $\checkmark$ & $\checkmark$ & $\checkmark$ \\
\midrule
\multicolumn{4}{l}{\textbf{Steering Control}} \\
\midrule
toward track center & $\checkmark$ & $\checkmark$ & $\checkmark$ \\
toward outside of curves &  & & $\checkmark$ \\
speed-based steering & $\checkmark$ &  &  \\
to counter angular velocity & & $\checkmark$ & $\checkmark$  \\
anticipate curvature & $\checkmark$ & $\checkmark$ & $\checkmark$ \\
\midrule
\multicolumn{4}{l}{\textbf{Specific Cases}} \\
\midrule
avoid grass & $\checkmark$ &  &  \\
avoid curbs & $\checkmark$ &  &  \\
recover when off-track & $\checkmark$ &  &  \\
brake when turning sharply & $\checkmark$ &  & $\checkmark$ \\
brake when track is lost & $\checkmark$ &  &  \\
brake when losing traction & & $\checkmark$ & $\checkmark$ \\
prevent full stop & $\checkmark$ &  &  \\
\midrule
\textbf{Total coverage} & 15 & 10 & 12 \\
\bottomrule
\end{tabular}
\caption{Strategy categories identified by manual inspection of Car Racing policies.}
\label{tab:semantic}
\end{table}

\subsection{Diagnostic categories and coverage}
\label{app:coverage}

The generated analyses connect these strategies to measurable properties of execution through four diagnostic categories.
Together they examine the values a policy computes, whether those values follow its definitions, how they relate to behavior, and how control signals evolve over time.

\paragraph{Summary statistics.}
Means, standard deviations, ranges, and signs characterize observations, semantic latents, and trainable parameters.
For example, the slip coefficient's sign tests whether its contribution to target speed agrees with the intended speed-reducing strategy.

\paragraph{Reconstruction errors.}
Analyses recompute a latent or action mean from its declared inputs and compare the result with the rollout value.
These checks test arithmetic consistency and help distinguish a mismatched computation from variation introduced by action sampling or logging.
In the slip example, accurate target-speed reconstruction establishes that the program executes the specified formula even while its learned coefficient violates the expected sign.

\paragraph{Correlations and regressions.}
Pearson correlations and linear regressions measure relationships between observations, latents, and actions across the trajectory.
For example, steering as a function of lateral offset tests whether the car tends to recenter itself, connecting a named strategy to its expression during execution.

\paragraph{Temporal behavior and action limits.}
Other diagnostics examine abrupt changes in control signals and the frequency of action saturation.
These checks identify behavior that may be obscured by aggregate return or trajectory-wide averages.

Table~\ref{tab:test-coverage} summarizes generated diagnostics across three refinement rounds, with 13, 10, and 10 test functions, respectively.
Their coverage changes as the policy evolves: later rounds add checks of angular velocity, off-track behavior, and steering at sharp turns.
This combination turns the semantic structure into an interface for testing and revision: each diagnostic identifies a policy quantity, a behavioral expectation, and a computation that can be edited.
Appendix~\ref{app:records} shows how these connections guide updates to features, parameter constraints, and state-dependent interactions.

\begin{table}[!t]
\small
\centering
\begin{tabular}{l|c|c|c}
\toprule
\textbf{Data Analysis} & \textbf{Iter 1} & \textbf{Iter 2} & \textbf{Iter 3} \\
\midrule
\multicolumn{4}{l}{\textbf{Sanity Checks}} \\
\midrule
Weights stay constant & $\checkmark$ & $\checkmark$ & $\checkmark$ \\
Signs consistent with assumptions & $\checkmark$ & $\checkmark$ & $\checkmark$ \\
Summary statistics & $\checkmark$ &   & \\
\midrule
\multicolumn{4}{l}{\textbf{Actions}} \\
\midrule
Output reconstruction & $\checkmark$ &  & $\checkmark$ \\
Latent reconstruction & $\checkmark$ & $\checkmark$ & $\checkmark$ \\
Action saturation & $\checkmark$ & $\checkmark$ & $\checkmark$ \\
Acceleration/braking consistency & $\checkmark$ & $\checkmark$ &  \\
Jerkiness & $\checkmark$ &  & $\checkmark$ \\
\midrule
\multicolumn{4}{l}{\textbf{Behavior}} \\
\midrule
Steering corrects lateral offset & $\checkmark$ &   & \\
Steering corrects heading error &  &   & \\
Slow at sharp turns & $\checkmark$ & $\checkmark$  & \\
Slow at edge of the track &  & $\checkmark$  & \\
Slow when angular velocity is high &  & $\checkmark$  & \\
Slow when wheels are slipping & $\checkmark$ & $\checkmark$  & \\
Off-track portion &  &  & $\checkmark$ \\
\midrule
\multicolumn{4}{l}{\textbf{Coefficient Importance}} \\
\midrule
Extra steering at sharp turns &  &  & $\checkmark$ \\
\midrule
\textbf{Number of test functions} & 13 & 10 & 10 \\
\bottomrule
\end{tabular}
\caption{Generated diagnostics adapt to the policy across three refinement rounds.
Checks cover computational consistency, action behavior, and task-specific control relationships.}
\label{tab:test-coverage}
\end{table}

\section{Policy refinement examples}
\label{app:records}

The following examples illustrate how generated diagnostics guide policy changes in Car Racing.

\subsection{Off-track recovery behavior}
\label{app:recovery-record}

The recovery behavior analysis filters rollout steps where the absolute lateral offset exceeds \texttt{half\_track\_width}, then checks whether steering points toward the track center under the recorded sign convention.
In the 1,000-step rollout, 74 steps are off-track and only 14.9\% of these have center-directed steering.
This identifies a state-dependent failure that aggregate return does not locate: the centering contribution is present, but rarely determines the steering direction when recovery is needed.
The revision adds an off-track indicator and a trainable nonnegative recovery strength, forming a multiplier that increases centering beyond the track boundary (Figure~\ref{fig:latent-traces}).
The other steering contributions remain in place.

\subsection{Car Racing programs, diagnostics, and rollouts}
\label{app:car-record}

Table~\ref{tab:matched-record} follows the slip-to-target-speed branch through three revisions.
Repeated sign checks lead from an unconstrained wheel-mean feature to a compressed positive residual with a nonpositive contribution to target speed.

\begin{table}[!t]
\centering\small
\begin{tabular}{@{}clrp{.40\linewidth}@{}}
\toprule
Round & Slip feature & Fitted $w_q$ & Diagnosis and revision \\
\midrule
0 & Mean wheel reading & $+0.1729$ & Positive coefficient violates the intended sign; replace wheel mean with a relative proxy. \\
1 & $\bar u-v$ & $+3.8367$ & Positive coefficient persists; rectify the residual. \\
2 & $\max(\bar u-v,0)$ & $+6.6077$ & Sign anomaly persists; constrain the coefficient to be nonpositive. \\
3 & $p/(1+\alpha p)$ & $-0.00390$ & Compression and $w_q=-|\eta|$ preserve the branch's speed-reducing role after fitting. \\
\bottomrule
\end{tabular}
\caption{Successive revisions align the slip proxy's learned contribution with its intended speed-reducing role.}
\label{tab:matched-record}
\end{table}

\begin{figure}[!t]
\centering
\begin{minipage}[t]{.49\linewidth}
\centering\includegraphics[width=\linewidth]{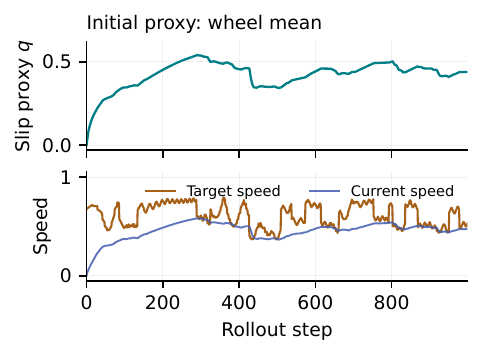}
\end{minipage}\hfill
\begin{minipage}[t]{.49\linewidth}
\centering\includegraphics[width=\linewidth]{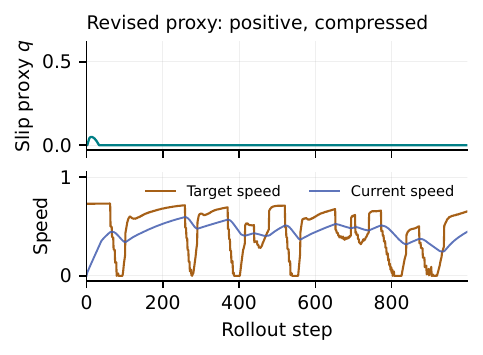}
\end{minipage}
\caption{Recorded slip and target-speed values before refinement (left) and after three revisions (right), with shared axis limits.}
\label{fig:slip-traces}
\end{figure}

Figure~\ref{fig:slip-traces} compares 1,000-step rollouts of the initial and third-round policies.
The shared axis limits expose changes in the slip proxy's scale and the resulting target-speed trajectories.

The slip example uses a generated sign check together with an arithmetic reconstruction check from the same round.
Listing~\ref{lst:test} shows the slip-coefficient check.

\nolinenumbers
\begin{minipage}{\linewidth}
\begin{lstlisting}[language=Python,caption={Abbreviated generated sign check from round 0.},label={lst:test}]
checks = {
    "weight_speed_slip": ("negative", lambda v: v < 0.0),
}
for col, (expect, fn) in checks.items():
    if col not in df.columns:
        _append(f"{col}: MISSING")
        continue
    mean_val = _safe_get(df, col).mean()
    ok = fn(mean_val)
    _append(f"{col}: mean={mean_val:.6f}, expected {expect} -> {'PASS' if ok else 'FAIL'}")
\end{lstlisting}
\end{minipage}
\linenumbers

The target-speed calculation satisfies the reconstruction check, but its positive slip coefficient fails the expected sign check (Listing~\ref{lst:results}).
\nolinenumbers
\begin{lstlisting}[caption={Target-speed reconstruction and slip-sign diagnostics at round 0.},label={lst:results}]
=== test_2_action_consistency ===
desired_speed reconstruction: mean_abs_diff=2.296791e-08, max_abs_diff=7.852650e-08, frac_mismatch=0.000

=== test_4_signs_and_bounds ===
weight_speed_slip: mean=0.172909, expected negative -> FAIL
\end{lstlisting}
\linenumbers
The generated explanation interprets the wheel mean as a speed-related signal and proposes comparing it with vehicle speed.
Later analyses revisit the coefficient after fitting the revised proxy, leading to rectification and an explicit sign constraint.

\section{Implementation details}
\label{app:training}

\subsection{Refinement procedure and execution}
\label{app:procedure}
Algorithm~\ref{alg:refinement} summarizes the data flow and permitted interaction branches.
Here $\Sigma(C)$ denotes the table-column schema exposed by the environment and the policy's logging interface.
Analysis-code generation depends on this schema and the policy program, independently of the numerical rollout values.

\begin{algorithm}[htbp]
\caption{Refinement through semantic rollout analysis}
\label{alg:refinement}
\begin{algorithmic}[1]
\REQUIRE Task specification $S$, demonstrations $\mathcal D$, environment
\STATE $C\gets\mathrm{LLM}(S,\text{generation specification})$
\WHILE{the refinement procedure continues}
    \STATE $\theta\gets\mathrm{Fit}(C,\mathcal D)$; $\hat\pi\gets\pi_{C,\theta}$
    \STATE $T\gets\mathrm{LLM}(S,C,\Sigma(C),\text{diagnostic prompt})$
    \STATE $X\gets\mathrm{RolloutLog}(C,\theta)$
    \REPEAT
        \STATE $r\gets\mathrm{Execute}(T,X)$
        \STATE $b\gets\mathrm{LLM}(S,C,T,r,\text{revision context})$
        \IF{$b$ requests additional diagnostics}
            \STATE $T\gets$ additional analysis code
        \ENDIF
    \UNTIL{$b$ proposes a revised program or stops refinement}
    \IF{$b$ stops refinement} \STATE \textbf{break} \ENDIF
    \STATE $C\gets$ revised program from $b$
\ENDWHILE
\STATE \textbf{return} $\hat\pi$; optionally apply post-refinement RL
\end{algorithmic}
\end{algorithm}

The refinement loop has a configurable iteration budget, with a default maximum of six rounds and early termination through the LLM's stop response.
Additional tests run on the current table before the next revision decision.
If fitting a revised program fails, the LLM can regenerate it using the preceding policy and diagnostic context, with up to five attempts.

\subsection{Post-refinement reinforcement learning}
\label{app:rl}

We optimize a clipped policy-gradient objective using group-relative reward-to-go and window likelihoods.
For trajectory $i$, let $T_i$ be its length and let $\mathcal I_t$ contain trajectories valid at time $t$.
We compute
\begin{equation}
 R_t^i=\sum_{u=t}^{T_i-1}\gamma^{u-t}r_u^i,
 \qquad A_t^i=\frac{R_t^i-\mu_t}{\sigma_t+10^{-8}},
 \label{eq:rl-advantage}
\end{equation}
where $\mu_t$ and $\sigma_t$ are the population mean and standard deviation over $\mathcal I_t$.

The likelihood ratio multiplies action likelihoods over a window of $L$ valid steps.
In log-probability form,
\begin{equation}
 \rho_{it}(\theta)=\exp\!\left[
 \sum_{u=t}^{t+L-1}\bigl(\log\pi_\theta(a_u^i\mid s_u^i)
 -\log\pi_{\theta_{\mathrm{old}}}(a_u^i\mid s_u^i)\bigr)\right].
 \label{eq:rl-ratio}
\end{equation}
For valid window starts, the clipped objective is
\begin{equation}
 \mathcal J(\theta)=\mathbb E_{i,t}\!\left[
 \min\!\left(\rho_{it}(\theta)A_t^i,
 \operatorname{clip}(\rho_{it}(\theta),1-\epsilon,1+\epsilon)A_t^i\right)\right].
 \label{eq:rl-objective}
\end{equation}
An entropy bonus encourages exploration, with its coefficient decaying linearly over training.
The optimizer defaults are $\gamma=0.98$, clipping threshold $\epsilon=0.2$, Adam learning rate $0.007$, 20 sampled episodes per update, three optimization epochs, and initial entropy coefficient 0.01; the default Car Racing likelihood window is 150 steps.
Unlike GSPO's length-normalized sequence ratio~\citep{zheng2025group}, Eq.~\eqref{eq:rl-ratio} uses the product of likelihoods within each window.
Trajectories share an initial condition and are grouped by time step for advantage estimation.

\section{Policy-generation prompts and interfaces}
\label{app:generation}

The prompts specify the interface between the LLM, the trainable policy, and the execution logger.

\subsection{Shared generation specification}
\label{app:specification}

The shared specification asks for meaningful intermediate computations, trainable coefficients, and an instrumentation dictionary.
It also specifies action-distribution outputs and initialization based on expected signs and scales.
\begin{Verbatim}
* The model should represent the causal structure of a good policy for the task. Each intermediate step in the policy should have semantic meanings and the learnable parameters should only connect the variables that are correlated (or have a causal relationship) as opposed to a generic MLP where all the hidden nodes are connected.
* For a discrete output space, the model should predict a (potentially unnormalized) distribution among those discrete actions. For a continuous action space, predict the parameters of a multivariate normal distribution for the action (action mean and the lower triangle matrix L such that Sigma = LL^T). Make sure the $L$ is a lower Cholesky matrix. If you think the actions should not be correlated then the L can be a diagonal matrix for simplicity.
* Along with the action distribution prediction, also return an `info_dict` which is a dictionary that records all the intermediate steps and the weights. This will be aggregated into a table for further analysis. For the weights, name the key `weight_{variable_name}` and for the intermediate steps, name the key `latent_{variable_name}`. If an intermediate feature has multiple dimensions, break it down to multiple keys. But keep the name semantically meaningful. For example, instead of `latent_1`, `latent_2`, ... use `latent_{feature_name}` instead. Each value in the dictionary should be of shape `(1,)` or `(B,)` where B is the batch size.
* Keep the model simple.
* Represent new features as linear combinations of old features when possible. If you are very certain that no bias term is needed (i.e., the feature value should be 0 if all inputs are 0), then don't include the bias term to make it easier to learn.
* There might be cases where a linear combination is not sufficient; then you may use other operations, such as multiplication, to represent the interaction between two features.
* Register all weights and bias terms as `nn.Parameter` with `required_grad=True`. Don't use `register_buffer`.
* Don't name a parameter using PyTorch keywords such as `self.half`.
* When initializing the values for each weight and bias, set a value based on positive or negative correlation (e.g., if the input feature is negatively correlated, then set its weight to -0.1) instead of using random initialization. Pay attention to the scale of the input features and output actions and set the magnitude of the numbers accordingly.
* There should be no magic number in the `forward` function. Constants like 0, 1, and 0.5 are fine.
* Make sure all operations are differentiable so that the parameters can be learned by gradient descent.
* You may use any of the functions defined in PyTorch (e.g., `torch.logical_and`, `torch.clamp`, `torch.abs`, `torch.square`, etc.).
* Refrain from using `torch.sigmoid` or `torch.tanh` to avoid the vanishing gradient problem.
* Make sure the gradient can flow back to the parameters. Avoid in-place operations (always use a new variable name instead) or constructing new tensors (always use `torch.stack` or `torch.cat` instead).
* Assume the inputs are batched.
# Make sure the model handles `torch.nan` gracefully.
\end{Verbatim}

\subsection{Environment descriptions and observation schemas}
\label{app:environment}

\begin{figure}[!t]
\centering
\includegraphics[height=1.25in]{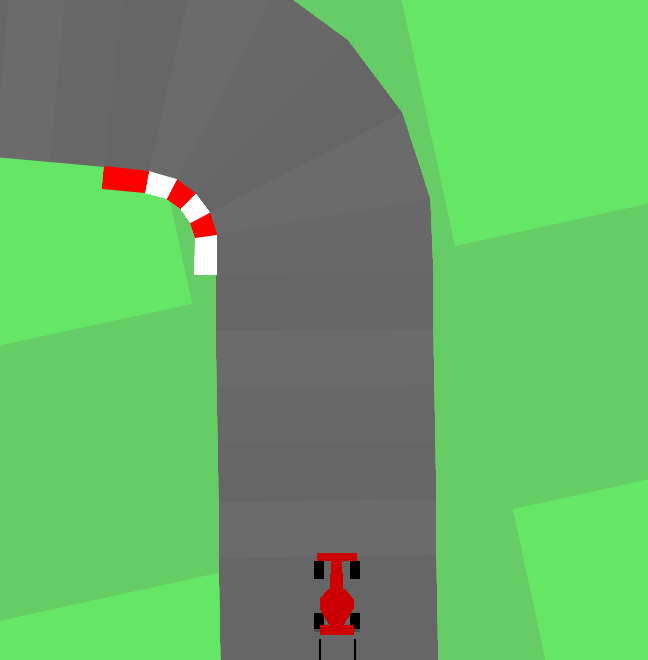}\hspace{0.55in}
\includegraphics[height=1.25in]{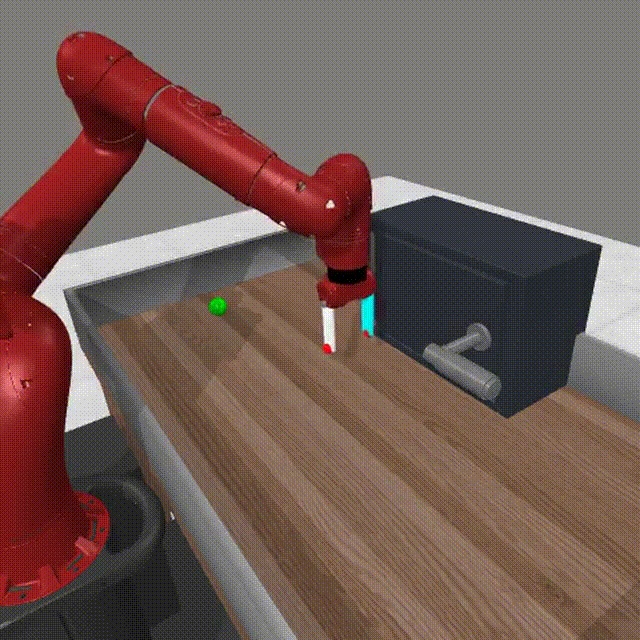}
\caption{Car Racing and Door Opening: continuous control from semantic observations.}
\label{fig:environments}
\end{figure}

Environment descriptions provide the semantic interpretation of each input and output coordinate.
Car Racing supplies track tiles and vehicle indicators; Door Opening supplies current and previous gripper/door information and a target position.
Figure~\ref{fig:environments} shows the two environments.

\paragraph{Car Racing.}
The following schema specifies the inputs to policy generation and diagnostic construction.
\begin{Verbatim}
# Features
The input to the model is two tensors representing the tracks and other indicator information:

0. (float32, (B, L, 4)) tiles. Where $B$ is the batch size and $L=12$ is the number of tiles on the track ahead of the race car, and the last dimension contains $(x, y, \theta, border)$ where $x$ is the signed lateral position of the tile (positive means the race car is on the left of the center of the tile), $y$ is the signed longitudinal position of the tile, $\theta$ is the relative heading of the tile to the race car (positive means the road is heading to the left), and $border$ is either 0 or 1, indicating whether the tile is a sharp corner (curvature exceeds some threshold).
1. (float32, (B, 7)) indicators. Where the first column is the speed $v$ (normalized from [0, 100] to [0, 1]), the second column is the current (absolute) heading of the race car $\theta$ (normalized from [-pi, pi] to [-1, 1]), the third column is the angular velocity $\omega$, and the rest are the abs sensor readings (approximate wheel omega) from each wheels (in the order of front left, front right, rear left, rear right).

# Output Space
A tensor of shape (B, 3) where each element corresponds to steer, accelerate, and brake controls of the race car, respectively. These controls are continuous. The steer value should be in the range of $(-1, 1)$, where -1 is full left, 0 is no steering, and 1 is full right. Accelerate is in the range of $(0, 1)$, where 0 is no acceleration and 1 is full acceleration. Brake is also in the range of $(0, 1)$, where 0 is no braking and 1 is full braking.

# Objective and Reward
The objective is to complete the track as fast as possible without going off the track. A positive reward is given whenever the race car moves to a new tile on the track and a small negative reward is given at each step as time penalty.

# Additional Notes
* The track is a loop.
* The track is about 0.4 unit wide. When the closest tile's x is at -0.2 or 0.2 it means the race car is on the edge of the track. Beyond that it's grass.
* The race car is approximately 3 tiles long.
* The first tile is the closest tile to the race car, and the rest are in the order of tiles ahead.
* The arguments of the forward function should only be self, tiles, and indicators.
* Estimate a variance for each output dimension to account for uncertainty.
* Clip the output to the desired range.
* Name your model "RacecarPolicy".
\end{Verbatim}

\paragraph{Door Opening.}
The description includes current and previous observations, allowing a generated policy to use information about recent motion.
\begin{Verbatim}
# Features
The input to the model is a tensor representing the position of the robot's gripper and the door at the current and previous time step.

Features: B x 39 where B is the batch size
0-2. The x, y, z coordinates of the gripper at the current time step.
3. A scalar value that represents how open/close the gripper is at the current time step.
4-6. The x, y, z coordinates of the door.
7-10. The quaternion (x, y, z, w) representing the orientation of the door.
11-17. Not used
18-20. The x, y, z coordinates of the gripper at the previous time step.
21. A scalar value that represents how open/close the gripper was at the previous time step.
22-24. The x, y, z coordinates of the door at the previous time step.
25-28. The quaternion (x, y, z, w) representing the orientation of the door at the previous time step.
29-35. Not used
36-38. The x, y, z coordinates of the target position for the door to be opened to.

# Output Space
A tensor of shape (B, 4) where the first three elements are the x, y, z motion of the gripper relative to its current position, and the last element is the gripper open/close command (1 for open, 0 for close).

# Objective and Reward
The objective is to reach for the handle on the door and pull the door open to a desired angle. The reward is given by the distance to the desired angle, when the gripper first hold onto the handle, and with a large positive reward for successfully opening the door. A small negative reward for each time step taken.

# Additional Notes
* The door revolves around a vertical axis.
* The door handle is horizontal.
* The gripper always points down.
* Name your model "OpenDoorPolicy".
\end{Verbatim}

\section{Diagnostic and revision prompts}
\label{app:prompts}

The diagnostic and revision prompts connect policy semantics to numerical execution evidence.
The LLM first proposes analyses using the code and schema, then receives their results as the basis for a possible edit.

\subsection{Hypothesis generation and analysis execution}
\label{app:diagnostic-prompts}

The system prompt asks the LLM to consider nominal behavior and possible anomalies before writing analysis code.
It requests explicit hypotheses, defensive functions over a dataframe, and output in a common results file.
The full numerical rollout is an input to these generated functions rather than part of the initial diagnostic-generation context.
\begin{Verbatim}
You are trying to find patterns in data and use it to enhance a task-specific PyTorch model that represents the causal structure of the policy. Specifically, you have access to a `pd.DataFrame` that records the model's intermediate steps as well as the observations from the environment during some policy rollouts. 

The model has a sparse, task-specific structure and its weights are trained using Imitation Learning from one demonstration.
More specifically, the model were created with the following specifications:
{SPECS}

You are going to do the following:

1) Analyze the task domain. List the important factors / behaviors of successfully completing the task. Think about what some suboptimal behaviors for the task may be in general, not just specifically for the given model structure.
2) Analyze the model definition and determine what it is for. Write a summary on what each weight and computation is used for, and how you can verify that the learned values are sensible. Essentially, you should form some hypotheses on the data by analyzing the structure of the model. Define what some expected nominal behavior is and what may be some anomalies for both the policy and the environment (e.g., as the policy is getting rolled out, how do you expect the observations to change).
3) For each expected behavior and potential anomaly, formulate some hypothesis that you can test given the rollout data. Make sure your tests cover all of the potential issues. Also, formulate some tests that will help you analyze the behavior of the policy. As the policy is trained with potentially incompatible demonstrations, it may perform very badly. So, you should verify if the policy demonstrates some of the suboptimal behaviors that you have reflected earlier.
4) Write scripts that help you process the data and test the hypotheses. You might find many time series feature processing techniques, data filtering, as well as statistical tests useful. These scripts should help you find the anomalies. Each of your functions should only take in `df` as the input, and should print out all the results to `test_results.txt`. Put all of your tests in a `run_all(df)` function. Remember to include all relevant imports. You may use `numpy, scipy, pandas, statsmodels, scikit-learn` among others, as well as custom helper functions. Keep your code defensive and try to handle errors gracefully. Once the scripts finish running, you can see the result of `test_results.txt`. Focus on the model design. Assume that there is no bug in the code execution and no missing values in the dataframe.
5) Ultimately, the results of these scripts should inform you of the problem of the current policy structure that allow you to propose updates to the original model structure (e.g., create a new latent variable, change to another operation, enforce a bound on the learned parameters, etc.) such that it performs better after another session of Imitation Learning training.

When you are writing the test scripts, keep in mind that:
* Action sampling is handled outside of model using `torch.distributions`. Specifically, discrete action spaces will be sampled with `Categorical(logits=action_probs).rsample()` and continuous action spaces will be sampled with `MultivariateNormal(action_mean, scale_tril=action_cov_tril).rsample()`.
* Actions will be automatically clipped to desired ranges after sampling. So it is normal for the final output action to have a different variance than the model output `action_cov_tril`.

Use the following return format:

[Task Analysis]

[Model Summary]

[Expected Nominal Behavior]

[Potential Anomalies and Tests]

[Hypotheses and Tests]

[Code Implementation]
```py
import ...

def test_1(df):
    """Brief description of test 1"""
    try:
        # actual test implementation
    except Exception as e:
        # log the error e into the same result file.

def run_all(df):
    ...
```
\end{Verbatim}

The per-policy template supplies the environment description, model definition, and available columns.
\begin{Verbatim}
[Environment Description]
{ENV_DESCRIPTION}

[Policy Structure]
```py
{POLICY_CODE}
```

[DataFrame Columns]
{DF_COLUMNS}

\end{Verbatim}

\subsection{Revision, additional tests, and termination}
\label{app:revision-prompts}

The revision prompt offers three responses: provide an updated program, request more diagnostic tests, or indicate that no further action is needed.
It allows changes to the computations and parameterization while preserving the model interface and logging outputs.
\begin{Verbatim}
test_results.txt
```text
{TEST_RESULTS}
```

Analyze the test results and compare them with the model summary, expected nominal behavior, and potential anomalies. Formulate some hypotheses on why the model may diverge from expected nominal behavior and think about what in the environment might be missing in the current structure. You may assume that the model already converged under Imitation Learning so that the weights already maximize the similarity to expert actions given the current structure.

Based on these analysis you can choose one of the following three options:

1) If you have a clear idea on how to improve the model, then explain the reasons and suggest the updates. Keep in mind that you are trying to design the model to learn better with Imitation Learning. You may add a new function like `torch.abs`, introduce a new latent variable, initialize the weights to different values, or add any differentiable operations. Keep the model simple to ensure learning stability. Remember not to edit variables in-place since that will confuse the gradient descent step. Keep the model explainable by prioritizing latent variables with semantic information as opposed to generic MLPs. No need to sample the actions when predicting the action mean, the sampling will be taken care of outside of the model. But you can control the sampling process by updating the output `scale_tril` matrix (e.g, adding upper/lower bounds, setting some variables to be independent, etc.) Include the description of the entire new model in the "Model Summary" section, not just the changes. Do not forget to return all weights and the intermediate steps (latent variables) in the return dictionary. Make sure each value in the `info` dict is either of shape (B,) or a scale. If there are multiple dimensions for a variable then break it down to multiple keys and use their physical meaning as keys (for example use "latent_action_steer" instead of "latent_action_0"). Do not forget to include all relevant imports and keep the function interface the same. Use the following format:

[Analysis of Test Results]

[Model Update Suggestions]

[Model Summary]

[Updated Model]
```py
import torch
from torch import nn
# other relevant imports

class ModelName(nn.Module):
    def __init__(self):
        ...

    def forward(self, tiles, indicators):
        ...
        return actions, cov_matrix, info
```

2) If you think the current test results are inconclusive, explain why and suggest additional tests or data that could help clarify the situation. Do not include the previously generated tests. Only include the additional tests. As a reminder you may only write to `test_results.txt` and should not create other files. Use the following format:

[Analysis of Test Results]

[Additional Tests]
```py
import ...

def additional_test_1(df):
    """Brief description of additional test 1"""
    ...

def run_all(df):
    ...
```

3) If you find no limitations and the model seems to be functioning well, confirm this using the following format:

[Analysis of Test Results]

[Done - No Further Action Needed]
\end{Verbatim}

\end{document}